\documentclass[sigconf]{acmart}

\AtBeginDocument{%
  \providecommand\BibTeX{{%
    \normalfont B\kern-0.5em{\scshape i\kern-0.25em b}\kern-0.8em\TeX}}}

\begin{document}

\title[Enhancing Bloodstain Pattern Analysis Through AI-Based Image Segmentation]{Enhancing Bloodstain Analysis Through AI-Based Segmentation: Leveraging Segment Anything Model for Crime Scene Investigation}

\author{Zihan Dong}
\email{zdong7@ncsu.edu}
\orcid{0000-0003-4079-7520}
\authornotemark[1]
\affiliation{%
  \institution{North Carolina State University}
  \streetaddress{890 Oval Drive}
  \city{Raleigh}
  \state{North Carolina}
  \country{USA}
  \postcode{27606}
}

\author{Zhengdong Zhang}
\email{zzhang3135@gatech.edu}
\orcid{0000-xxxx-xxxx-xxxx}
\affiliation{%
 \institution{Georgia Institute of Technology}
 \streetaddress{890 Oval Drive}
 \city{Cary}
 \state{North Carolina}
 \country{USA}}
 \postcode{27519}

\renewcommand{\shortauthors}{Dong and Zhang}

\begin{abstract}
    Bloodstain pattern analysis plays a crucial role in crime scene investigations by providing valuable information through the study of unique blood patterns. Conventional image analysis methods, like Thresholding and Contrast, impose stringent requirements on the image background and is labor-intensive in the context of droplet image segmentation. The Segment Anything Model (SAM), a recently proposed method for extensive image recognition, is yet to be adequately assessed for its accuracy and efficiency on bloodstain image segmentation. This paper explores the application of pre-trained SAM and fine-tuned SAM on bloodstain image segmentation with diverse image backgrounds. Experiment results indicate that both pre-trained and fine-tuned SAM perform the bloodstain image segmentation task with satisfactory accuracy and efficiency, while fine-tuned SAM achieves an overall 2.2\% accuracy improvement than pre-trained SAM and 4.70\% acceleration in terms of speed for image recognition. Analysis of factors that influence bloodstain recognition is carried out. This research demonstrates the potential application of SAM on bloodstain image segmentation, showcasing the effectiveness of Artificial Intelligence application in criminology research. We release all code and demos at \url{https://github.com/Zdong104/Bloodstain_Analysis_Ai_Tool}.   
\end{abstract}




\received{14 June 2023}
\received[revised]{12 June 2023}
\received[accepted]{5 June 2023}

\maketitle

\section{Introduction}
The significance of bloodstain pattern analysis lies in its ability to provide crucial information about a crime scene by studying the unique patterns formed by blood, revealing details about what might have occurred\cite{Singh07}. Traditionally, this analysis has been conducted manually, requiring extensive expertise and time-consuming efforts. However, recent advancements in computer vision and artificial intelligence (AI) technology have innovated solutions that can revolutionize this process.

One such solution is the Segment Anything Model (SAM)\cite{kirillov2023segment} developed by Meta. This innovative model has the impressive capability of accurately and efficiently segmenting various objects and elements in images, with a combination of points, boxes, and text as the input prompt. Such a breakthrough technology has the potential of fueling a myriad of downstream tasks, particularly bloodstain prediction in this paper, which benefits both mechanical engineers and criminology analysts.

This paper represents the initial investigation into the application of SAM for blood droplet prediction in the context of bloodstain pattern analysis to offer engineers and criminology analysts the ability to achieve accurate and efficient bloodstain image segmentation. A fine-tuned SAM with accurately segment bloodstain dataset is trained. Comparison of Thresholding, Thresholding with Median Filtering \cite{bhargavi2014survey, 1163188}, pre-trained SAM, and fine-tuned SAM is conducted to determine the performance of model in accuracy, efficiency on bloodstain segmentation.  This study contributes to enhanced data efficiency and optimized utilization of hardware resources, providing convenience and valuable insights for crime scene analysis.

\begin{figure}[h]
  \centering
  \includegraphics[width=0.9\linewidth]{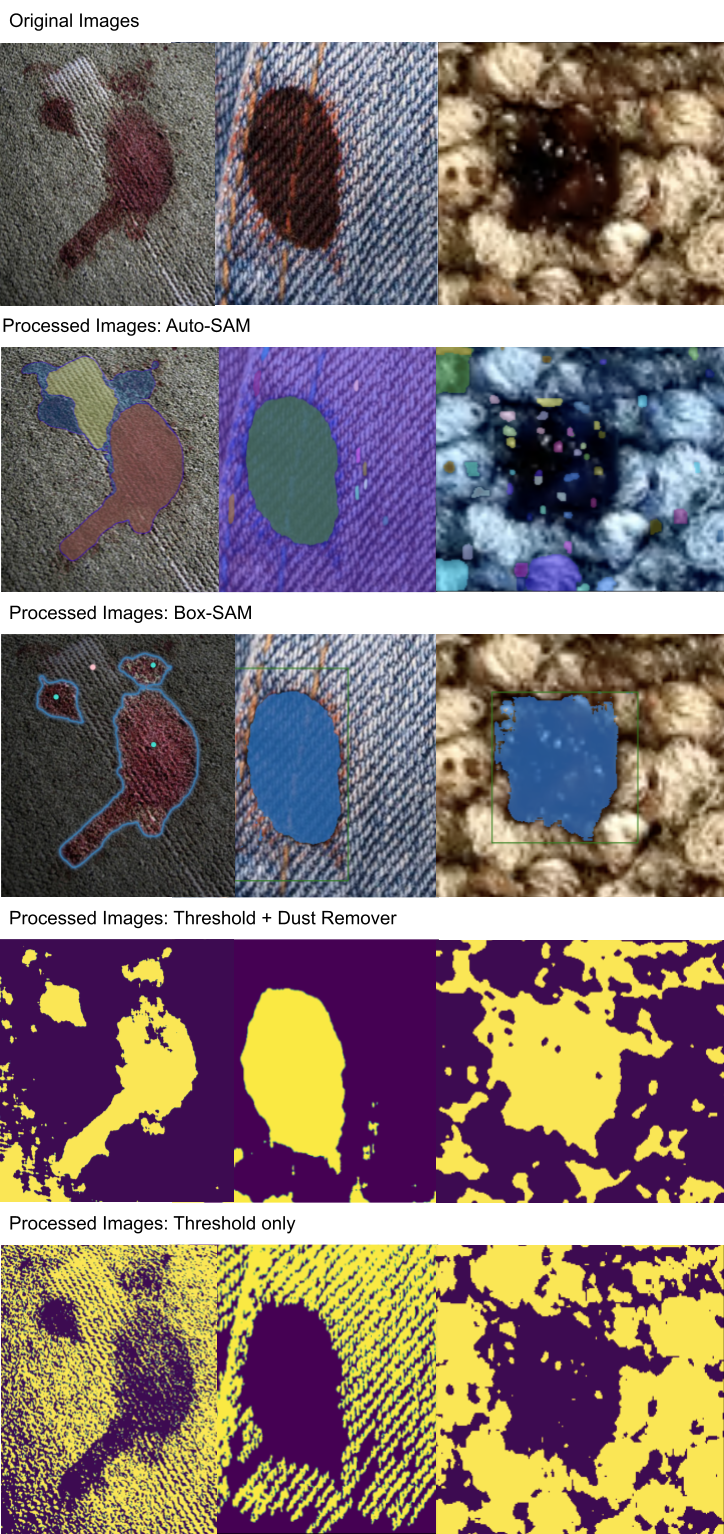}
  \caption{Test for different conditions and methods, on left: carpet, middle: jean and right: wool}
  \Description{The SAM model is much better than the conventional thresholding method}
  \label{fig:test_conditions}
\end{figure}

\section{Existing Work}
\subsection{Work on Bloodstain Pattern Analysis}
Bloodstain pattern analysis (BPA) has been employed as a means of crime scene reconstruction and testimonial evidence for more than one hundred years. \cite{james2005principles}. BPA interprets the bloodstain patterns in a crime scene in order to provide evidence to support the crime scene reconstruction. \cite{liu2020automatic} In recent years, there is a growing trend in forensic science to develop methods to make forensic pattern comparison tasks more objective. This has generally involved the application of suitable image-processing methods to provide numerical data for identification or comparison. \cite{arthur2017image} Thresholding is an image processing technique used to split an image into smaller segments, or junks, using at least one color or gray scale value to define their boundary. \cite{bhargavi2014survey} In the field of BPA, Thresholding is commonly used to segment bloodstain from blood spatter pattern. Traditional thresholding-based methods using tools like ImageJ/Fiji\cite{Image_J} and Matlab are still widely used today. Thresholding involves setting a threshold value to distinguish objects from the background based on intensity variation. However, choosing the right threshold is critical since different parts of an image may require different intensity variations to accurately capture the edges of interest \cite{PAL19931277,41371}. Besides, these methods are constrained by their reliance on images with monochrome backgrounds and strong contrast, such as white backgrounds and red/brown blood droplets. Moreover, variations in camera angle, appearance, and the presence of other objects can decrease detection accuracy and pose challenges for bloodstain pattern analysis \cite{mirani2022object}.

\subsection{Work on AI-based Image Segmentation}

\begin{figure}[h]
  \centering
  \includegraphics[width=\linewidth]{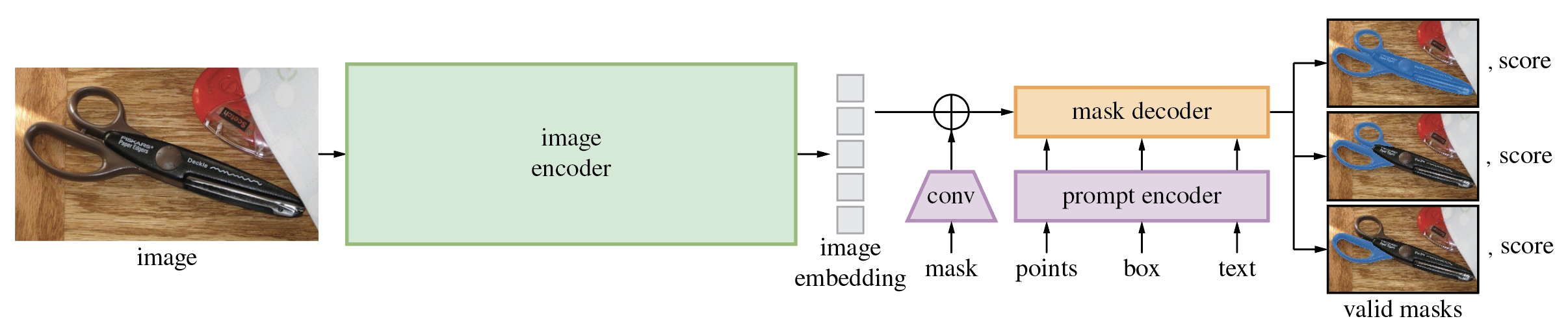}
  \caption{SAM model diagram \cite{kirillov2023segment}}
  \Description{this diagram shows what is the mechanism of SAM model}
  \label{fig:SAM_Model}
\end{figure}

 With the emergence of AI, there has been a surge in innovative methods to image segmentation, mainly through various machine learning and deep learning algorithms. Early works in AI-based image segmentation typically involves traditional machine learning algorithms such as the k-means clustering algorithm\cite{luo2003spatial}. Later on, the introduction of Convolutional Neural Networks (CNNs) \cite{milletari2016v} improved efficiency and accuracy in image segmentation tasks, with further enhancements achieved through the use of region-based CNNs \cite{wei2021automatic}. 

Recently, Meta AI Research launched Segment Anything Model (SAM), which aims to be a foundation model for image segmentation. \cite{kirillov2023segment} Unlike traditional methods, SAM is promptable. It can generate accurate segmentation masks based on different prompts. These prompts can take the form of foreground/background points, rough boxes or masks, free-form text, or any other input that specifies the object to be segmented in an image. (Figure \ref{fig:SAM_Model}) This flexibility allows SAM to generalize to unfamiliar objects and images without compromising its segmentation capabilities.

Existing research on the combination of AI and bloodstain pattern analysis primarily focuses on tutorial-based approaches, teaching professors how to use K Nearest Neighbor(KNN) for bloodstain drip analysis and blood droplet pattern classification and categorization. \cite{ravivarma2021development} However, there is very limited existing work on applying AI-based tools to bloodstain segmentation, especially in realistic scenarios with diverse backgrounds. In this paper, we explore the performance of pre-trained and fine-tuned SAM with different prompts on bloodstain segmentation.

In the field of mechanical engineering, the skills required for data analysis and Python programming are often not covered in the curriculum. Most of the widely accepted AI models and new technologies are written in Python, making it challenging for engineers to adopt them. Additionally, tools like Anaconda and Python installations, as well as module installations through pip, are not widely known. Many engineers still rely on proprietary software like Matlab, which is not open source. This limited access to the latest AI models online leads to decreased working efficiency.

\section{Method}
\subsection{Data}
It is challenging to acquire clean and high-resolution images necessary for fluid dynamic analysis of bloodstains. From a public available high speed digital videos set, Midwest Forensic Resource Centre (MFRC) Blood Pattern Analysis Videos created by National Institute of Justice (NIJ), we have selected 18 videos filmed from various of impact angles for images and masks generation. These videos serve as valuable resources for studying drip bloodstains on different impact surfaces. \cite{MFRC}. 

Frames from high-speed digital videos are captured as image sets and selected to be used in the training set for fine-tuning SAM. Those images are processed with the threshold method \cite{yanowitz1989new} to obtain ground truth masks. Additional manual adjustments and data verification were performed using Fiji, an image processing tool to enhance the accuracy of those masks. The final dataset, prepared for model fine-tuning, consists of 400 images and their corresponding ground truth masks, covering blood droplets dripped from various angles (10 degrees, 20 degrees, 30 degrees, 40 degrees, 50 degrees, 60 degrees, 70 degrees, 80 degrees, and 90 degrees). RGB images are also converted to gray-scales for additional image sets to reinforce the model performance on grey-scale images, which is very common in high-speed digital video collection and recognition. The additional randomly-chosen images selected from other videos in the MFRC set are employed for testing and evaluating the accuracy of the fine-tuned SAM, pre-trained SAM, Thresholding and Thresholding with Median Filtering.

\subsection{Model}
In order to strike a balance between high efficiency and accuracy, this paper selects the smallest variant of the Segment Anything Model (SAM), vit-b model, for fine-tuning. This choice is made to ensure efficient processing of bloodstain segmentation while maintaining a high level of accuracy. By opting for the vit-b model, the computational requirements and processing time is minimized, making it well-suited for practical implementation.

\subsection{Model Fine-tuning Process}
We improve the performance of the pre-trained vit-b model on segmenting bloodstains through model fine-tuning, leveraging the 400 MFRC images of drip bloodstains and their ground truth masks. The wide range of drip angles aptly encapsulates most instances the fine-tuned model is projected to encounter in real-world applications.

The model fine-tuning is performed by comparing predicted masks to ground truth masks and updating the model's parameters based on the computed loss. The key steps involved in this process are as follows:

\begin{itemize}
\item {First, the input images and ground truth masks are manipulated to match the format and dimensions that SAM anticipates. } 
\item {Second, each bloodstain image is marked with bounding box prompt. Given that each image contains only one blood droplet, this is conveniently accomplished by designating the full dimensions of the image as a bounding box. } 
\item {Following this, 10-epoch training process is conducted to fine-tune the mode. In each epoch, the mean squared error (MSE) is calculated by comparing the predicted binary mask to the ground truth binary mask.}
\item {After computing the loss, the optimizer's gradients are reset to zero, and the loss is backpropagated to compute the gradients for all the model's parameters. The optimizer then takes a step to update the parameters based on the computed gradients.}
\end{itemize}

\section{Performance of Models}
In this section, we evaluate the performance of Segment Anything Models with different kinds of prompts and modes. Testing dataset selected is different from the training set with the same image size (230x700). The definition of model names used in this paper are defined as below:

\begin{itemize}
    \item Auto-Default: the pre-trained SAM that generates masks with a large number of single-point input prompts over bloodstain images.
    \item Auto-Fine Tuned: the fine-tuned SAM that generates masks with a large number of single-point input prompts over bloodstain images.
    \item Box-Default: the pre-trained SAM that generates masks with a single bounding box on the bloodstain image as input prompt.
    \item Box-Fine Tuned: the fine-tuned SAM that generates masks with a single bounding box on bloodstain images as input prompt.
    \item Point-Default: the pre-trained SAM that generates masks with a single point on bloodstain images as input prompt.
    \item Point-Fine Tuned: the fine-tuned SAM that generates masks with a single point on bloodstain images as input prompt.
\end{itemize}

\begin{figure}[h]
  \centering
  \includegraphics[width=0.8\linewidth]{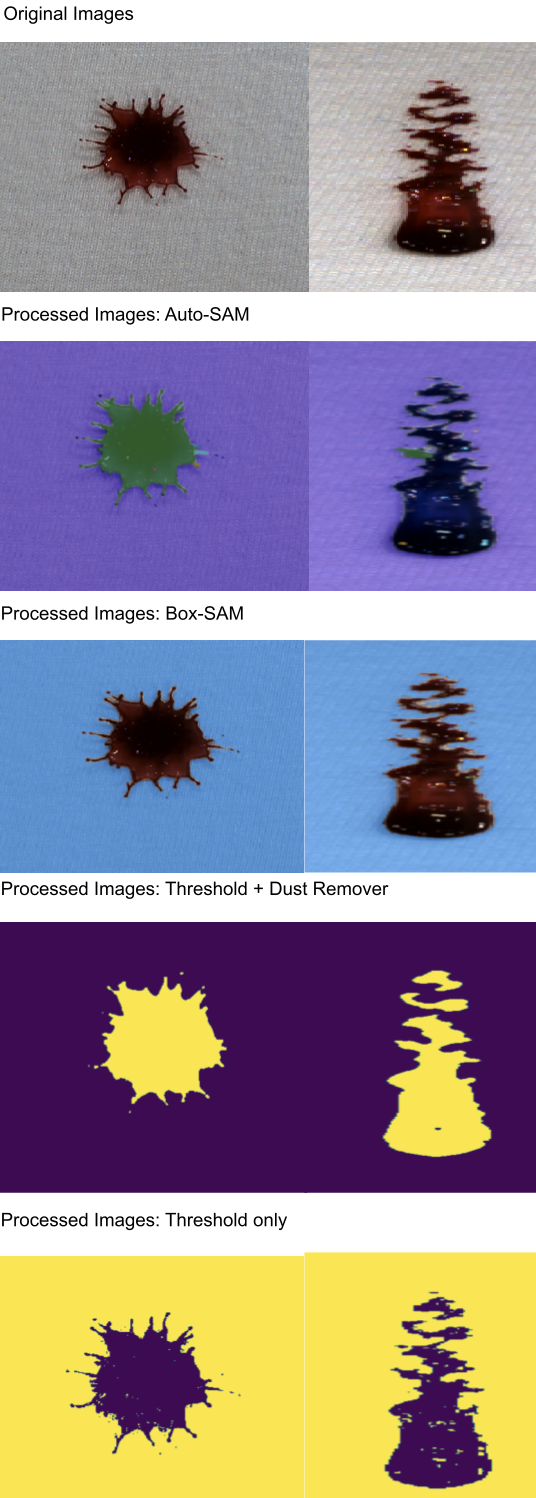}
  \caption{Method comparison for SAMs and Threholds}
  \Description{The figure compared the original images, Auto SAM, Box SAM, Point SAM, Threshold with Median Blur and Threshold only.}
  \label{fig:Method Comparison}
\end{figure}

\subsection{Accuracy}
The model accuracy is accessed by determining whether predicted masks show considerable lack of recognition in the bloodstain segmentation. For example, shadows of blood droplet on the fabric should be clearly differentiated from bloodstain and not included in the segmented droplet.

The accuracy of pre-trained and fine-tuned SAM is evaluated on a testing dataset of 180 images that consist of 90 RGB images and 90 greyscale images. The testing dataset is composed of a series of 10 images that were randomly selected in 10-degree increments from 10 to 90 degree. To obtain a comprehensive accuracy evaluation, we tested SAMs under different types of prompt: Automatic SAM (zero-shot), Box Prompt, and Point prompts. (See Table \ref{tab: Model with Diff})

The results of the model accuracy evaluation are as follows:

When applied to gray-scale bloodstain image using Auto mode, the pre-trained SAM exhibits a slight inferior performance than on the RGB images. 

For the pre-trained model, the segmentation accuracy is 97.8\% under bounding box prompt and point prompt. However, the Auto mode has accuracy of 87.8\% (79/90) recognition among the images.

For the fine-tuned model, the accuracy for bloodstain segmentation with bounding box prompt is 97.8\% and point prompt is 98.9\%, and the accuracy of Auto mode decreases to 91.1\%.

In the case of the pre-trained SAM with gray-scale images, the Auto mode achieved an accuracy of 87.8\%, and points and bounding box prompts achieved 97.8\% in accuracy.

After fine-tuning, the auto-SAM model achieved an accuracy of 90.00\% (81/90) for points and bounding box prompts, with an bloodstain image segmentation overall accuracy of 91.1\% (82/90).

\begin{table*}
  \caption{Model with Different Mode and Accuracy }
  \label{tab: Model with Diff}
  \begin{tabular}{cccl}
    \toprule
    Mode \& Model & RGB Accuracy(90) & Grey Accuracy(90) & Overall Accuracy(180) \\
    \midrule
    Auto-Default & 87.8 & 87.8 & 87.8 \\
    Auto-Fine Tuned & 92.2 & 90.0 & 91.1 \\
    Box-Default & 97.8 & 97.8 & 97.8 \\
    Box-Fine Tuned & 97.8 & 97.8 & 97.8 \\
    Point-Default & 97.8 & 97.8 & 97.8 \\
    Point-Fine Tuned & 98.9 & 98.9 & 98.9 \\
    \bottomrule
  \end{tabular}
\end{table*}

In summary, the fine-tuned SAM model demonstrated improved accuracy compared to the pre-trained model, and the use of traditional methods and prompts helped ensure the validity of the results. Challenges such as capillarity-induced branches, shadow effects, and low contrast were identified and potential solutions were suggested.

\subsection{Efficiency}
To evaluate efficiency, the T4 GPU was used as the standard compute power for the SAM model accelerator. The time required for manual adjustments, comparisons, and benchmark of each method is added to this standard compute power to determine the time taken by each method (Table \ref{tab: Image_processing}).

Efficiency plays a crucial role in evaluating the performance of Segment Anything models. Through discriminative fine-tuning of the entire network, considerable speedups can be achieved with minimal loss in accuracy, as demonstrated in the fine-tuned SAM \cite{lebedev2014speeding}.

The default SAM exhibits reasonable processing time for average-sized images containing a single blood droplet completing processing in approximately 2.5 seconds using a single T4 GPU. Similarly, the fine-tuned SAM demonstrated improved efficiency, with an average processing time of 2.39 seconds for the same type of image. On average, the processing speed has 4.7\% improvement with the fine tune process for original model. These processing times ensure that the models can be effectively utilized within practical time constraints. 

Table \ref{tab: Image_processing} presents the time required for image processing using the NVIDIA T4 GPU on Google Colab. 

\begin{table}
  \caption{Image processing time on NVIDIA T4 GPU}
  \label{tab: Image_processing}
  \begin{tabular}{ccl}
    \toprule
    Index & RGB (seconds)\\
    \midrule
    SAM-Default & 2.5 \\
    SAM-Fine Tuned & 2.39 \\
    Percentage Faster & 4.70\% \\
  \bottomrule
  \label{Table:Processing Time}
\end{tabular}
\end{table}

\subsection{Shot Number}
Shot-number is another important aspect to consider when evaluating the efficiency of the SAM models. In this case, the box and point provided in the image remain constant for every image in the dataset. The size of the box is chosen to be the same as the image size, and the point is consistently placed at the center of the image.

The advantage of using a constant box and point configuration is that it eliminates the need for manual adjustment or intervention during the processing of large datasets. By keeping the box and point placement uniform across all images, the models can process the dataset faster and more efficiently. This ensures a streamlined and automated workflow without any human interruptions.

The use of a fixed shot-number methodology reduces variability and allows for quicker analysis of the dataset as a whole. It minimizes the need for manual intervention and ensures a consistent approach to image processing. This approach contributes to the overall efficiency of the SAM models, enabling faster and more reliable results for large datasets.

\subsection{Discussion}
Efficiency in image processing is influenced by various factors, including the lighting conditions during filming and the video angle. Adequate lighting plays a critical role in enhancing contrast and facilitating accurate image recognition. Stronger contrast leads to clearer boundaries between objects, enabling the model to distinguish and segment them more effectively. Moreover, capturing images at a 90-degree angle, or as close to it as possible, is highly recommended to achieve optimal results. This angle provides a comprehensive view of the droplets, minimizing distortions and occlusions caused by oblique angles. By ensuring favorable lighting conditions and appropriate camera angles, the overall efficiency and accuracy of the SAM models can be significantly improved.

Table \ref{tab: Factors_Affecting} outlines the factors that can affect blood droplet images and their corresponding influences. Factors such as filming angle, image color, filming brightness, and background color can all have an impact on the accuracy of the model's predictions.

\begin{table}
  \caption{Factors Affecting Blood Droplet Images}
  \label{tab: Factors_Affecting}
  \begin{tabular}{cl}
    \toprule
    Factors & Influence\\
    \midrule
    Filming Angle & 0.1\\
    Image Color & 0.5\\
    Filming Brightness & 0.3\\
    Background Color & 0.1\\
  \bottomrule
\end{tabular}
\label{Factors effect result}
\end{table}

In addition to the statistical data analysis provided above, we also have the following innovative discovery:
\begin{itemize}
    \item Small branches in the image caused by capillarity were ignored due to the small size of the segmentation, but this issue could be addressed by increasing the image size and quality for additional analysis.
    \item Detecting the shape of dropping objects proved to be challenging due to the presence of shadows, and recognizing two-color images was also difficult due to the shadow effect.rates masks with a single bounding box on the bloodstain image as input prompt.
    \item It is worth noting that higher contrast in images can make them more striking and recognizable to both humans and the models. \cite{High_Contrast}  The presence of strong contrast aids in enhancing image features, making them easier to identify and classify accurately. This correlation between image quality, contrast, and model performance highlights the importance of capturing high-quality images with optimal contrast levels to achieve better accuracy and explainability in the SAM models. The low contrast of some images made them harder to recognize. To improve recognition accuracy, stronger contrast lighting during filming is recommended.
    \item When background is clean and in monochrome, threshold has a overall superior performance than SAM in efficiency, However, the SAM shows a much stronger segmentation performance when the bloodstain image captured has low resolution, background include a large amount of variations.
    \item With additional test on different test set from the selected dataset, we find the RGB images need a longer processing time than grayscale images due to the increase in image size will cause a longer time for image processing.
\end{itemize}

The accuracy of these models is influenced by various factors, such as image quality and the background of the processed images. In the specific test case described, the images feature a clean white background, which enables clear and precise recognition of the fabric texture. As a result, both box and point predictions can achieve accuracy levels of up to 100\%. 

The auto-segmentation model utilized in this scenario is trained using a dataset that includes multiple masks. Consequently, there may be instances where the model associates small changes in images with other objects, potentially leading to inaccuracies. However, adjusting the default auto-segmentation parameters to different values can address such issues and improve overall accuracy.

In cases where there is only one object present in the image, and the multi-mask option is set to False for box and point predictions, the accuracy of the models can be further enhanced. By focusing solely on a single object, the models can provide more precise and reliable predictions.

When it comes to RGB images, they tend to exhibit higher accuracy in image prediction compared to gray-scale or other color spaces. The additional color channels in RGB images provide more information for the models to analyze, leading to improved recognition and prediction capabilities.

\subsection{Traditional Method vs SAM for Segmentation}
To evaluate the accuracy and efficiency of the traditional threshold method and the AI-based segmentation method (SAM), a test was conducted on a dataset of 180 images. The evaluation form mentioned earlier in this section was utilized for accuracy assessment. The results of this comparison shed light on the limitations of the threshold method in object detection, as the threshold fails to accurately detect the blood droplets (objects) and often includes the background as part of the segmented image. This inefficiency makes it challenging for researchers to extract the blood droplets and conduct further experiments. Furthermore, determining the appropriate threshold parameter to ensure accurate object detection while excluding background noise proves to be a difficult task.

By comparing the segmentation results of the Thresholding method (Thres), the Thresholding method combined with the Median Filtering technique (Thres + MF), default SAM model with boxed prompt for whole image and the fine tuned SAM with boxed prompt for whole image on the 180 test samples, we obtained the accuracy percentages listed in Table \ref{tab: Accuracy_for_Thes_v_Sam }. Although the inclusion of Median Filtering in the threshold method led to a significant improvement in image segmentation, it still resulted in information loss and decreased segmentation accuracy. To generate a reliable segmentation, adjusting parameters is required for each videos and it makes the data processing step inefficient.

On the other hand, the fine tuned SAM exhibited a notable accuracy of 87.8\%, outperforming both the threshold method and the threshold method with Median Filtering, and demonstrated good segmentation performance. In addition, the SAM does not adjust of parameters for each individual videos which convenience the data processing. The results clearly demonstrate the superiority of the SAM model in terms of accuracy and efficiency for bloodstain segmentation.

\begin{table}
  \caption{Accuracy for Threshold vs. SAM on 180 images}
  \label{tab: Accuracy_for_Thes_v_Sam }
  \begin{tabular}{cccl}
    \toprule
    Thres& Thres + MF & DB & FB\\
    \midrule
    74.4\% & 81.1\% & 97.8\% & 97.8\% \\
  \bottomrule
  \label{Accuracy Threshold SAM, Thres is Threshold, MB: Median Blur, DB: Default Box, FB: Fine Tuned Box}
\end{tabular}
\end{table}

\section{Limitations}
The utilization of large images in the experimental setup necessitates relatively high computational power from the GPU, in contrast to the traditional threshold method that primarily relies on CPU power, resulting in longer processing times and more compute power. This becomes particularly challenging in real-world scenarios where GPU power is not as prevalent as CPU power.

To assess the performance of the threshold method, a series of tests were conducted on diverse images. It was observed that the threshold method exhibits superior overall performance, taking into account resource efficiency and time efficiency, when the images possess a pristine background and high-quality resolution.

During the process of fine-tuning the model, our dataset solely comprised of individual droplet analyses from various angles. Consequently, the model's efficacy in analyzing other types of bloodstains, such as bloodstain spray or large bloodstain areas, cannot be guaranteed.

\section{Boarder Impact}
In addition to its application in bloodstain pattern analysis, our model also exhibits reliable performance in predicting various types of droplets. The extensive training, involving 1 million images and 11 billion segmentation data, has endowed the fine tuned SAM \cite{kirillov2023segment} with robust recognition capabilities for diverse objects. Notably, the similarity between droplet shapes and blood droplet shapes contributes to the model's effectiveness in this specific domain. Furthermore, the material composition of the droplets has minimal influence on computer vision-based image processing.

Beyond the scope of this paper, we have provided a GitHub notebook demo that allows users to import the model to Google Colab for further fine-tuning, tailored to their specific requirements. Leveraging the already potent object detection and segmentation abilities of SAM, the fine-tuning process necessitates only a small training set.

By offering this notebook demo, we aim to empower researchers and practitioners to utilize the model's strong capabilities for their specific applications. The accessibility and adaptability of the model open up possibilities for a wide range of computer vision tasks beyond bloodstain analysis, facilitating advancements in various fields.

\section{Conclusion and Future Works}
Based on the research conducted, it is evident that the SAM model outperforms threshold-based methods in terms of overall performance, especially when processing large datasets. The AI-generated results have demonstrated higher accuracy and reduced information loss compared to threshold methods or a combination of Thresholding and Median Filtering techniques. This is particularly beneficial for analyzing detailed and important images.

Moving forward, future works could involve the integration of mechanical knowledge and fluid dynamics principles to develop a comprehensive system for blood droplet detection. By combining AI model predictions with domain-specific knowledge, such a system can provide valuable output parameters. Furthermore, there is potential for further optimization of the fine-tuned model to enhance data efficiency, model efficiency, and hardware efficiency, thereby reducing the compute budget.

Further advancements can involve the development of even smaller models that can be deployed on personal laptops, expanding accessibility to a wider range of users. Additionally, the creation of efficient software tools specifically designed for engineers and professionals in the field of criminology analysis who do not have very strong coding skills would be highly valuable.

In summary, the research conducted has demonstrated the superiority of the fine-tuned SAM model over traditional threshold-based methods. Future works can focus on the integration of domain knowledge, optimization of the model, and the development of user-friendly software to enhance efficiency and expand the applicability of the system to various domains and user groups.

\begin{acks}
We express our gratitude to Zhiyuan for providing valuable assistance in reviewing and providing suggestions for this paper. We would also like to extend our thanks to Peng for his contribution in reviewing and offering insightful suggestions for improvement.

Furthermore, we would like to acknowledge the indispensable guidance provided by Dr. Dongkuan throughout the process of structuring this paper. His expertise and instructions have been instrumental in shaping the framework of this research work.

We gratefully acknowledge the referees for their thoughtful comments that have helped to improve the paper significantly.

We sincerely appreciate the support and contributions of all those mentioned above, as their input has significantly enhanced the quality and coherence of this academic paper.
\end{acks}

\section{Online Resources}
We release all code and demos at \url{https://github.com/Zdong104/Bloodstain_Analysis_Ai_Tool}.  We highly recommend researchers to run this code on Google Colaboratory which provides free GPU and strong computational power. 

\bibliographystyle{ACM-Reference-Format} 
\bibliography{references} 

\end{document}